\documentclass[sigconf]{acmart}

\usepackage{natbib}
\usepackage{amsfonts}       
\usepackage{nicefrac}       
\usepackage{microtype}      
\usepackage{multirow}
\usepackage{subcaption}
\usepackage{pifont}
\usepackage{makecell}
\usepackage{float}
\usepackage{enumitem}
\usepackage{colortbl}
\usepackage{inputenc} 
\usepackage{fontenc} 
\usepackage{url} 

\newcommand{\cmark}{\ding{51}}
\newcommand{\xmark}{\ding{55}}

\AtBeginDocument{%
  }

\copyrightyear{2024}
\acmYear{2024}
\setcopyright{acmlicensed}\acmConference[MMASIA '24]{ACM Multimedia Asia}{December 3--6, 2024}{Auckland, New Zealand}
\acmBooktitle{ACM Multimedia Asia (MMASIA '24), December 3--6, 2024, Auckland, New Zealand}
\acmDOI{10.1145/3696409.3700217}
\acmISBN{979-8-4007-1273-9/24/12}

\begin{document}

\title{Collaborative Feature-Logits Contrastive Learning for Open-Set Semi-Supervised Object Detection}







\author{Xinhao Zhong}
\affiliation{
    \institution{Institute of Information Science, Beijing Jiaotong University, Beijing Key Laboratory of Advanced Information Science and Network}
    \city{Beijing}
    \country{China}
}
\author{Siyu Jiao}
\affiliation{
    \institution{Institute of Information Science, Beijing Jiaotong University, Beijing Key Laboratory of Advanced Information Science and Network}
    \city{Beijing}
    \country{China}
}
\author{Yao Zhao}
\affiliation{
    \institution{Institute of Information Science, Beijing Jiaotong University, Beijing Key Laboratory of Advanced Information Science and Network}
    \city{Beijing}
    \country{China}
}
\author{Yunchao Wei$^{\dag}$}
\thanks{\dag~Corresponding author.}
\affiliation{
    \institution{Institute of Information Science, Beijing Jiaotong University, Beijing Key Laboratory of Advanced Information Science and Network}
    \city{Beijing}
    \country{China}
}

\renewcommand{\shortauthors}{Xinhao Zhong et al.}

\begin{abstract}
  Current Semi-Supervised Object Detection (SSOD) methods enhance detector performance by leveraging large amounts of unlabeled data, assuming that both labeled and unlabeled data share the same label space. However, in open-set scenarios, the unlabeled dataset contains both in-distribution (ID) classes and out-of-distribution (OOD) classes. Applying semi-supervised detectors in such settings can lead to misclassifying OOD class as ID classes. To alleviate this issue, we propose a simple yet effective method, termed Collaborative Feature-Logits Detector (CFL-Detector). Specifically, we introduce a feature-level clustering method using contrastive loss to clarify vector boundaries in the feature space and highlight class differences. Additionally, by optimizing the logits-level uncertainty classification loss, the model enhances its ability to effectively distinguish between ID and OOD classes. Extensive experiments demonstrate that our method achieves state-of-the-art performance compared to existing methods.
\end{abstract}


\begin{teaserfigure}
  \includegraphics[width=0.9\linewidth]{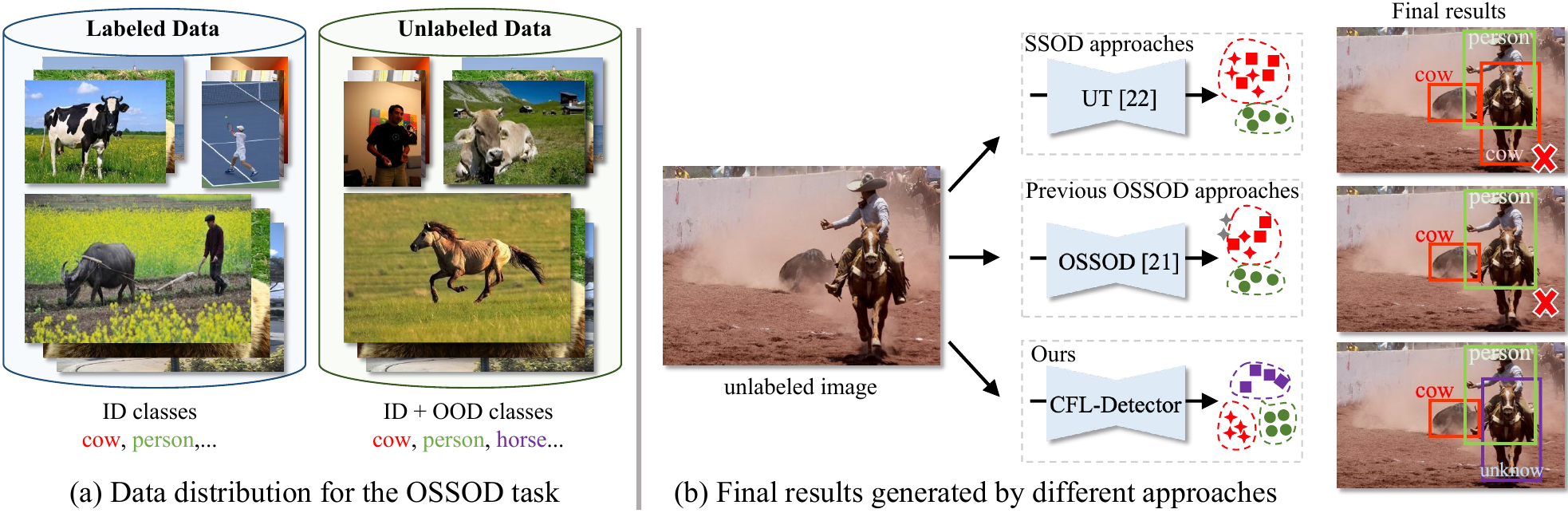}
  \centering
  \caption{(a) The dataset configuration for the OSSOD task includes both labeled and unlabeled data with different distributions. (b) We present visualizations comparing predictions from SSOD methods, previous OSSOD approaches, and our proposed method. The SSOD method UT \cite{liu2021unbiased} misidentifies "horse" (OOD class) as "cow" (ID class). Previous OSSOD approaches \cite{ossod} fail to detect "horse". Our CFL-Detector optimizes inter-class distributions, effectively detecting both ID classes ("cow", "person") and OOD class ("horse"). Throughout this paper, predictions for OOD class are uniformly labeled as "unknown".}  
  \Description{This is an image that shows a introduction of this paper.}
  \label{fig:intro}
\end{teaserfigure}


\maketitle

\section{Introduction}
\label{sec:intro}
\begin{figure*}[t]
\begin{center}
   \includegraphics[width=0.95\linewidth]{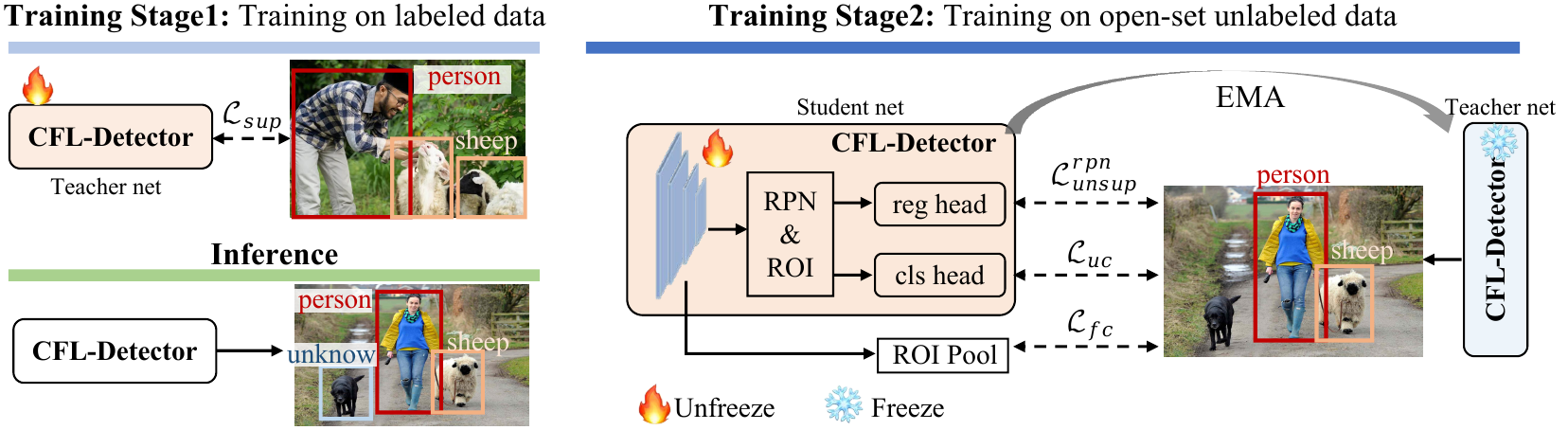}
\end{center}
   \vspace{-1mm}
   \caption{
    Overview of our CFL-Detector. Training Stage 1: We begin with fully supervised pre-training on labeled data, where the pre-trained model serves as the teacher net. Training Stage 2: The teacher net's parameters are frozen, and the student net is trained on unlabeled data using pseudo-labels from the teacher net. To ensure reliable pseudo-labels, the teacher net is updated with an exponential moving average (EMA) of the student net's parameters. Inference: our CFL-Detector effectively detects both ID classes ("person" and "sheep") and OOD class ("dog").
    \Description{This is an image that shows an overview of this paper.}
   }
\label{fig:overview}
\vspace{-1mm}
\end{figure*}
In recent years, semi-supervised learning (SSL) has demonstrated the importance of utilizing unlabeled data to enhance model performance. Initially focused on image classification \cite{xie2020self,sohn2020fixmatch}, SSL research later expanded to dense prediction tasks like object detection, which involves instance-level multi-object recognition. Inspired by SSL, several SSOD approaches \cite{tang2016large,yang2021interactive,tang2021proposal,tang2021humble,fang2023locating} have emerged. However, these methods are typically trained in close-set settings, making them ineffective in open scenarios. As illustrated in Fig. \ref{fig:intro} (a), the training dataset consists of a small amount of labeled data (containing only ID classes) and a large amount of unlabeled data (containing both ID and OOD classes). Consequently, open-set semi-supervised object detection (OSSOD) is proposed to tackle the challenges of detecting objects in open-world scenes.

As shown in Fig. \ref{fig:intro} (b), transferring SSOD methods directly to open-set settings is inappropriate. Due to the distribution mixture in the feature space, SSOD method like Unbiased Teacher (UT) \cite{liu2021unbiased} misclassifies OOD class as ID classes. For example, UT may classify a "horse" as a "cow". In contrast, our method can accurately identify OOD class (unknown). Previous work \cite{ossod} attempted to address this misclassification by training an offline OOD detector using labeled data to filter OOD instances from pseudo-labels, thereby reducing OOD interference. While this method improves ID class accuracy, it has limitations. Simply filtering out OOD instances cannot teach the model to distinguish OOD class, \textit{e.g.}, \cite{ossod} mistakenly filters a "horse" as "background", removing its bounding box in Fig. \ref{fig:intro}(b). Moreover, fine-tuning an OOD detector is time-consuming.

To address these issues, we propose the \textbf{C}ollaborative \textbf{F}eature-\textbf{L}ogits \textbf{Detector} (CFL-Detector), which effectively detects OOD class while preserving ID detection capability (Fig. \ref{fig:intro}). Specifically, we introduce feature contrastive loss ($\mathcal{L}_{fc}$) and uncertainty classification loss ($\mathcal{L}_{uc}$) to optimize distributions at both the feature and logits levels. $\mathcal{L}_{fc}$ operates at the feature level, clustering same-class features and separating those of different classes to make OOD features distinct. $\mathcal{L}_{uc}$ operates at the logits level, optimizing the classification loss for ID classes and the weighted classification loss for OOD class, enabling the model to learn probabilities for both.

Extensive experiments and ablation studies demonstrate that our method achieves state-of-the-art performance and can be effectively applied to other SSOD methods. Our contributions are as follows:

\begin{itemize}
    \item At the feature level, we introduce feature contrastive loss to reduce the distance between same-class features and increase the distance between different-class features.
    \item At the logits level, we introduce uncertainty classification loss to help the model learn the uncertainty of OOD class.
    \item The two components comprise our OSSOD framework (CFL-Detector), which is collaboratively trained and easily applicable to other SSOD methods.
\end{itemize}

\section{Related Work}
\label{sec:related}
\subsection{Open-Set Learning} 
Extensive research has explored open scenarios. Continual learning \cite{zhang2023slca,zhu2023ctp} enables learning new knowledge without forgetting prior knowledge. Few-shot learning \cite{zhang2021few,zhang2022mask,li2020meta} addresses recognizing novel categories with limited labeled data. Open-Vocabulary learning \cite{jiao2023learning,jiao2024collaborative,han2023global} classifies objects using arbitrary categories described by texts. In contrast, open-set learning focuses on detecting and distinguishing unknown categories from known ones. Early works, such as scherer et al. \cite{scheirer2014probability}, introduced a linear SVM classifier for open-set recognition. As research expanded into open-set object detection, OWOD \cite{joseph2021towards} proposed energy-based class differentiation, while OpenDet \cite{han2022expanding} analyzed latent space distributions. Since OOD classes can easily be confused with ID classes, many methods focus on maximizing OOD probability. However, optimizing probability alone is insufficient. Our method combines feature space analysis with uncertainty probabilities to enhance class discrimination.

\subsection{Semi-Supervised Object Detection} 
Semi-supervised learning enhances model performance by leveraging labeled and unlabeled data, reducing reliance on fully labeled datasets compared to weakly supervised approaches \cite{zhang2023credible,jiang2021online}. Recent advancements in semi-supervised object detection (SSOD) have shown substantial promise, particularly through methods such as data augmentation \cite{wang2021data,li2022pseco}, consistency regularization \cite{berthelot2019mixmatch,jeong2019consistency,zheng2022simmatch,zhang2023controlvideo}, and pseudo-labeling techniques \cite{zhou2021instant,li2022rethinking,liu2021unbiased}. Consistency-based methods enforce consistency constraints to utilize unlabeled data, while pseudo-label-based methods generate pseudo-labels to guide training. However, these methods are limited to close-set scenarios, as the presence of OOD class can degrade pseudo-label quality, leading to semantic confusion and reduced performance.

\subsection{Open-Set Semi-Supervised Learning} 
Open-set semi-supervised learning \cite{li2023iomatch,guo2020safe,cao2021open} presents significant challenges. OpenMatch \cite{saito2021openmatch} identified that OOD samples can adversely impact results, proposing a method to identify and filter such samples using auxiliary models. Inspired by \cite{saito2021openmatch}, the subsequent open-set semi-supervised techniques further improved this method to better distinguish OOD. \cite{ossod}, the pioneering work introducing the OSSOD task, follows a similar approach by fine-tuning a model to detect OOD class and filter misclassified bounding boxes offline. While these methods effectively reduce error information, they require additional model fine-tuning. Wang et al. \cite{wang2023online} proposed a threshold-free Dual Competing OOD head to mitigate the problem of error accumulation in semi-supervised outlier filtering. However, previous methods primarily focused on improving the accuracy of ID classes. To better reflect real-world scenarios, it is essential for models to also identify OOD class effectively.

\section{Preliminary}
\label{sec:prelimiary}
\subsection{Task Defination}
\label{sec:Task Defination}
The objective of Open-Set Semi-Supervised Object Detection (OSSOD) is to train a model using a limited amount of labeled ID data while effectively detecting OOD class during testing. The main challenge is to leverage a large amount of noisy, unlabeled data to help the model to differentiate between various classes. Formally, we denote the labeled dataset as $\mathcal{D}_l = \left \{ X_l, Y_l \right \} $ and the unlabeled dataset as $\mathcal{D}_u = \left \{ X_u \right \}$. The ID classes in $\mathcal{D}_l$ are represented as $C_k = \left \{ 1, 2, ..., K \right \}$, while the OOD class is uniformly represented as $C_{u} = K + 1$, with the background denoted as $C_{bg}$.

\subsection{Revisiting Unbiased Teacher Pipeline} 
\label{sec:Revisiting Unbiased Teacher pipeline}
The goal of Semi-Supervised Object Detection (SSOD) is to improve object detectors by leveraging a small labeled dataset $\mathcal{D}_{l}$ alongside a large unlabeled dataset $\mathcal{D}_{u}$. Unbiased Teacher (UT) \cite{liu2021unbiased} uses a teacher-student framework for self-training. Initially, the teacher net is trained on labeled data to gain basic detection capabilities. In the next stage, the teacher processes weakly augmented unlabeled images, applying confidence thresholds and Non-Maximum Suppression (NMS) to generate high-quality pseudo-labels. The student then uses strongly augmented images with these pseudo-labels for training. The loss function is defined as:
\begin{equation}
\mathcal{L}_{ssod} = \mathcal{L}_{s}(x_{s}, y_{s}) + \lambda \mathcal{L}_{u}(x_{u}, \hat y_{u})
\end{equation}
where $\hat y_{u}$ are pseudo-labels, and $\lambda$ is the unsupervised loss weight. The teacher's weights are updated using the exponential moving average (EMA) method \cite{tarvainen2017mean}. Although UT performs well in closed-set settings, its effectiveness diminishes in open-set scenarios.

\section{Method}
\label{sec:method}
\subsection{Overview}
\label{sec:Overview}
\begin{figure*}
\begin{center}
   \includegraphics[width=0.9\linewidth]{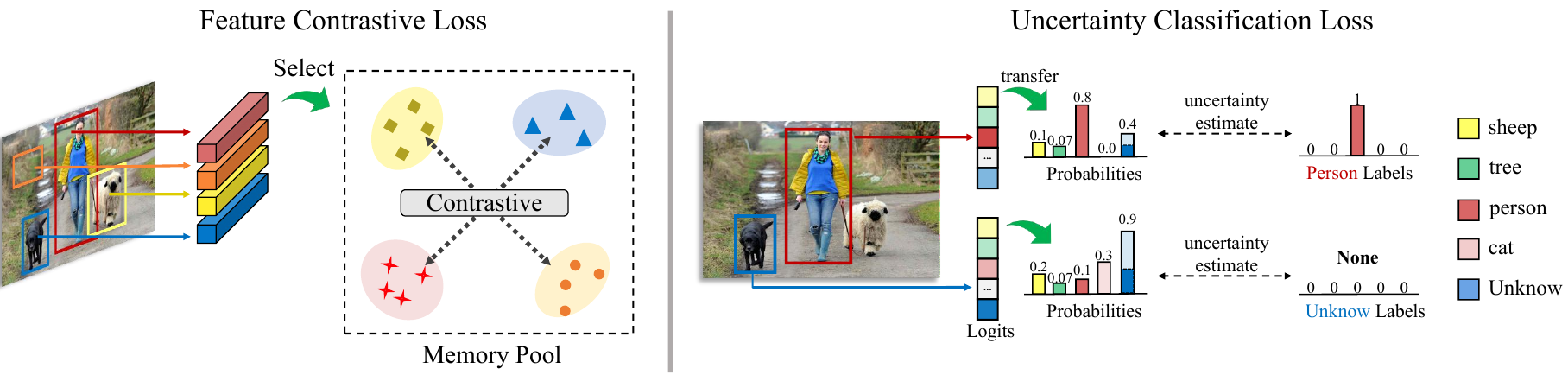}
\end{center}
\vspace{-1mm}
   \caption{
    Details of the feature contrastive loss ($\mathcal{L}_{fc}$) and uncertainty classification loss ($\mathcal{L}_{uc}$). $\mathbf{Left}$: The $\mathcal{L}_{fc}$ is computed over the feature pool, aggregating ID classes while separating OOD class. Thresholds continuously update the feature pool to maintain both diversity and accuracy. $\mathbf{Right}$: The $\mathcal{L}_{uc}$ modifies the cross-entropy calculation to assign higher confidence to the OOD, boosting the classifier’s ability to distinguish OOD.
   }
   \Description{This is an image that shows details of this paper.}
\label{fig:details}
\vspace{-1mm}
\end{figure*}

We introduce the CFL-Detector, which learns class distributions at both \textbf{feature} and \textbf{logits} levels. It uses ResNet for feature extraction and a detector for object detection (Sec. \ref{sec:Feature Extraction and Detector}). We introduce feature contrastive loss $\mathcal{L}_{fc}$ (Sec. \ref{sec:Feature Contrastive Loss}) and uncertainty classification loss $\mathcal{L}_{uc}$ (Sec. \ref{sec:Uncertainty Classification Loss}) to model class distributions. Our approach follows a two-stage training pipeline. Firstly, we conduct fully supervised training on labeled data to establish detection capabilities. Secondly, we utilize unlabeled data for semi-supervised training. The complete diagram of the CFL-Detector is shown in Fig. \ref{fig:overview}.

\subsection{Feature Extraction and Object Detection}
\label{sec:Feature Extraction and Detector}
\noindent \textbf{Feature Extraction.} We utilize a pre-trained ResNet50 \cite{he2016deep} to extract features from the input image $I$. The output of each stage of ResNet is denoted as $F=\{F_i\}, i \in [0,1,2,3]$. $F_{0}$, $F_{1}$, $F_{2}$, $F_{3}$ have strides of \{4, 8, 16, 32\} with respect to the input image. 

\noindent \textbf{Object Detection.} We employ Faster-RCNN \cite{faster2015towards} with a Feature Pyramid Network (FPN) \cite{lin2017feature} as our object detector. After extracting feature maps $F$ from the input image, these maps are fed into the Region Proposal Network (RPN) to generate region proposals. The detector then predicts the bounding box coordinates $b$ and the classification logits $s$ based on these proposals. Following post-processing, feature vectors $R = \{R_n\}, n = \{1, 2, ..., N\}$ are extracted from the feature pyramid through RoIAlign.

\noindent \textbf{Objective.} We map ${R_n}$ to a low-dimensional embedding $e_{n} \in \mathbb{R}^{d}$ (with $d$ = 128 by default) via a multi-layer perceptron (MLP). This embedding $e_{n}$ is used to compute the feature contrastive loss (Sec. \ref{sec:Feature Contrastive Loss}). To enable the model to learn the detection of OOD class, we utilize the classifier's logits to compute the uncertainty classification loss (Sec. \ref{sec:Uncertainty Classification Loss}). The total loss in our two-stage training is defined as:
\begin{equation}
    \mathcal{L} = \mathcal{L}_{sup} + \lambda \mathcal{L}_{unsup}
\end{equation}
where $\lambda $ is the weight for the unsupervised loss. The unsupervised and supervised losses are defined as:
\begin{equation}
\renewcommand{\arraystretch}{1.5}
\begin{array}{c}
    \mathcal{L}_{unsup} = \mathcal{L}_{cls}^{rpn} + \alpha_{t}\mathcal{L}_{fc} + \beta \mathcal{L}_{uc} \\
    \mathcal{L}_{sup} = \mathcal{L}_{cls}^{rpn} + \mathcal{L}_{reg}^{rpn} + \mathcal{L}_{reg}^{roi} + \alpha_{t}\mathcal{L}_{fc} + \beta \mathcal{L}_{uc} 
\end{array}
\end{equation}
where $\mathcal{L}_{cls}$ represents the classification loss and $\mathcal{L}_{reg}$ represents the regression loss, $\alpha_{t}$ and $\beta$ are hyper-parameters. Due to the low quality of the pseudo bounding box coordinates, we do not optimize the unsupervised regression loss.

\subsection{Feature Contrastive Loss}
\label{sec:Feature Contrastive Loss}
The contrastive loss aims to learn the similarities and differences between data samples, bringing similar pairs together while separating dissimilar pairs. Inspired by the supervised contrastive loss \cite{khosla2020supervised}, we adapt it for \textbf{feature-level} optimization.

As illustrated in Fig. \ref{fig:details} (left), we implement the feature contrastive loss ($\mathcal{L}_{fc}$) on embedding $e_{n}$. The objective is to minimize the distance between same-class embeddings while maximizing the distance between different-class embeddings. We introduce a memory pool $E = \{E_c\}$, where $c = \left \{ 1, 2, ..., K, K+1 \right \}$, to store embeddings of each class within the given image. To ensure the embedding quality in $E$, we implement an embedding filtering mechanism. For each embedding $e_n$, we compute the IoU score ($S_{IoU}$) between the bounding box coordinates and the ground truth, as well as the cosine similarity ($S_{cos}$) between the embedding and the class center. We retain only those embeddings, where $S_{IoU}$ > 0.7 and $S_{cos}$ > 0.5. Considering training costs, we do not store all embeddings that meet these criteria. Instead, we iteratively replace low-scoring embeddings with high-scoring ones, maintaining a maximum of $q = 256$ embeddings for each class in the memory pool.

Since we additionally store features predicted as "unknown" into the memory pool, these features carry diverse semantic information, which could confuse the model. To mitigate this issue, we do not compute intra-class similarity for OOD class. Instead, we concentrate on inter-class similarity and optimize proposals with $IoU > 0.5$ only. The $\mathcal{L}_{fc}$ is formulated as follows:
\begin{equation}
\renewcommand{\arraystretch}{1.5}
\begin{array}{c}
\mathcal{L}_{fc} = \frac{1}{N} \sum_{i=1}^{N} \mathcal{L}_{fc}(e_{i}) \\
\mathcal{L}_{fc}(e_{i}) = \frac{1}{{|K(c_{i})|}}\sum_{e_{k}\in K(c_{i})}log\frac{exp(e_{i}\cdot e_{k}/\tau )}{\sum_{e_{u}\in U(c_{i})}exp(e_{i}\cdot e_{u}/\tau )} 
\end{array}
\end{equation}
\noindent where $N$ denotes the number of proposals, $e_{i}$ is the $i$-th proposal embedding, $\tau$ is a temperature hyper-parameter, $K(c_{i})$ denotes the features of $c_{i} \in ID$ in the memory pool, and $U(c_{i})$ represents the features excluding $c_{i}$. 

\subsection{Uncertainty Classification Loss}
\label{sec:Uncertainty Classification Loss}

While the optimization of the feature contrastive loss helps in delineating boundaries between different classes at the feature level, the model still confuse the \textit{OOD class} with the \textit{ID class} under the cross-entropy loss due to the absence of annotations for OOD class. Inspired by Opendet \cite{han2022expanding}, we propose an uncertainty classification loss ($\mathcal{L}_{uc}$) that operates at the \textbf{logits-level} to help the model learn representations for the \textit{OOD class}.

As illustrated in Fig. \ref{fig:details} (right), within $\mathcal{L}_{uc}$, the $\mathrm{Softmax}$ function is employed to compute probabilities for \textit{ID class}. For \textit{OOD class}, if the model fails to directly identify, we mine potential OOD instances by selecting the top K candidates from the background class probabilities. We consider only the scores of the \textit{OOD class} and \textit{background class}, thus reinforcing the constraint on the OOD class. The probability of class $c$ ($p_{c}$) is defined as:
\begin{equation}
    p_{c} = \frac{\exp(s_{c})}{\sum _{j \in C} \exp(s_{j})}, \quad C = \left\{
    \begin{alignedat}{2}
    & C_{bg} \cup C_{u} \cup C_{k}, & \quad & \text{if } c = {ID} \\
    & C_{bg} \cup C_{u},        & \quad & \text{if } c = {OOD} \\
    \end{alignedat}
    \right.
\end{equation}
where $C$ denotes the class set, and we use uncertainty estimation to assign weights to each class. The higher the probability of the $C_k$, the lower its uncertainty. The formula of $w_c$ is defined as:
\begin{equation}
    w_{c} = \left\{
    \begin{alignedat}{2}
    & 1,        & \quad & \text{if } c = {ID} \\
    & (1 - p_{k})^{\alpha } p_{k}, & \quad & \text{if } c = {OOD} \\
    \end{alignedat}
    \right.
\end{equation}
where $p_{k}$ denotes the probabilities of $C_k$, and $\alpha$ is a hyper-parameter controlling the weight value. We compute $\mathcal{L}_{uc}$ as:
\begin{equation}
\label{eq:7}
    \mathcal L_{uc} = \left\{
    \begin{alignedat}{2}
    & -\sum w_u log(p_u) - \sum w_klog(p_k), & \quad & \text{if } sup \\
    & -\sum w_u log(p_u),        & \quad & \text{if } semi \\
    \end{alignedat}
    \right.
\end{equation}
where $w_u$ and $w_k$ denote the weights assigned to OOD and ID classes, respectively. The terms $sup$ and $semi$ refer to the supervised and semi-supervised stages. Our $\mathcal{L}_{uc}$ enhances the classifier's ability to distinguish between classes by assigning higher prediction probabilities to certain classes and lower prediction probabilities to others. This loss function (Eq. \ref{eq:7}) is essentially the sum of the weighted cross-entropy loss ($\mathcal{L}_{ce}$).

\section{Experiments}
\label{sec:experiments}
\subsection{Settings} 
\begin{table*}[t]
\caption{Comparison with state-of-the-art methods under different settings.}
\label{tab:coco-open}
\vspace{-2mm}
\centering
\centering
    \subfloat[Comparisons on the \textbf{COCO-Open-CLS} with 4000 labeled images.
    \label{tab:coco-open-1}]
    { 
    \renewcommand\tabcolsep{5.5pt}
    \renewcommand\arraystretch{1.1} 
    \small
    \begin{tabular}{l|ll|ll|ll}
      \Xhline{0.7pt}
      \centering
      \multirow{2}{*}{Num of classes} & \multicolumn{2}{c|}{20/60} & \multicolumn{2}{c|}{40/40} & \multicolumn{2}{c}{60/20} \\
      & mAP$_{k}$ & AP$_{u}$ & mAP$_{k}$ & AP$_{u}$ & mAP$_{k}$ & AP$_{u}$ \\
      \hline
      Label & 9.25 & 0 & 15.36 & 0 & 18.62 & 0 \\
      STAC \cite{sohn2020simple} & 10.44 & 0 & 17.47 & 0 & 22.51 & 0 \\
      UT \cite{liu2021unbiased} & 10.75 & 0 & 18.91 & 0 & 23.74  & 0 \\
      ORE \cite{joseph2021towards} & 9.66  & 1.74 & 16.03  & 2.41& 20.63& 4.21  \\ 
      Opendet \cite{han2022expanding} & 9.87  & 3.10  & 16.83  & 4.55 & 21.59  & 7.22\\
      \rowcolor{gray!20}
      \textbf{Ours} & \textbf{14.71} & \textbf{5.56} & \textbf{20.84} & \textbf{7.11} & \textbf{25.05}  & \textbf{9.82}  \\
      \Xhline{0.7pt}
      \end{tabular}
    }
\centering
    \subfloat[Comparisons on the \textbf{COCO-Open-supp} with 20 ID classes.
    \label{tab:coco-open-2}]
    { 
    \centering%
    \small 
    \renewcommand\tabcolsep{5.5pt}
    \renewcommand\arraystretch{1.1} 

    \begin{tabular}{l|ll|ll|ll}
      \Xhline{0.7pt}
      \centering
      \multirow{2}{*}{Num of Images} & \multicolumn{2}{c|}{1000} & \multicolumn{2}{c|}{2000} & \multicolumn{2}{c}{4000} \\
      & mAP$_{k}$ & AP$_{u}$ & mAP$_{k}$ & AP$_{u}$ & mAP$_{k}$ & AP$_{u}$ \\
      \hline
      Label & 5.11 & 0 & 6.72 & 0 & 9.45 & 0 \\
      STAC \cite{sohn2020simple} & 6.03 & 0 & 8.25  & 0 & 10.52  & 0 \\
      UT \cite{liu2021unbiased} & 6.51 & 0 & 8.63 & 0 & 10.98  & 0 \\
      ORE \cite{joseph2021towards} & 5.71 & 0.67  & 7.70 & 0.98  & 9.88  & 1.48  \\ 
      Opendet \cite{han2022expanding} & 5.86 & 1.46 & 7.81 & 2.21 & 10.07 & 2.99 \\
      \rowcolor{gray!20}
      \textbf{Ours} & \textbf{9.01} & \textbf{4.25} & \textbf{10.98} & \textbf{5.02} & \textbf{14.88} & \textbf{5.31} \\
      \Xhline{0.7pt}
      \end{tabular}
    } 

    
\end{table*}

\begin{table}[h]
\vspace{-2mm}
\caption{Comparisons on the \textbf{VOC-COCO}.}
\label{tab:voc-coco}

\centering%
\small 
\renewcommand\tabcolsep{5.5pt}
\renewcommand\arraystretch{1.1} 
    \begin{tabular}{c|cc|cc}
    \Xhline{0.7pt}
    Methods & Labeled & Unlabeled & mAP$_{k}$ & AP$_{u}$ \\
    \hline
    UT \cite{liu2021unbiased}  & VOC & COCO & 28.82 & 0 \\
    DCO OSSOD \cite{wang2023online} & VOC & COCO & 30.29 & 0 \\
    \rowcolor{gray!20}
    \textbf{Ours} & VOC & COCO & \textbf{31.49} & \textbf{7.86} \\
    \Xhline{0.7pt}
    \end{tabular}
\end{table}

\noindent \textbf{Dataset.} We adhere to the experimental protocols established in previous works \cite{ossod,wang2023online}, conducting evaluations on two popular OSSOD benchmarks, Pascal-VOC \cite{everingham2010pascal} and MS-COCO \cite{coco}. Our method is assessed under three distinct experimental settings: COCO-Open-CLS, COCO-Open-SUP, and VOC-COCO.

\begin{itemize}[itemsep=2pt,topsep=0pt,parsep=0pt]
    \item \textbf{COCO-Open-CLS.} We randomly select 20/40/60 classes as \textit{ID classes}, with the remaining classes as \textit{OOD class} in the MS-COCO dataset. 
    \item \textbf{COCO-Open-SUP.} We randomly select 1/2/4k images from MS-COCO, containing only objects from \textit{ID classes} as the labeled dataset. The remaining images, with annotations removed, serve as the unlabeled dataset. 
    \item \textbf{VOC-COCO.} The Pascal-VOC training set, containing 11,000 images and 20 classes, is used as labeled data. Meanwhile, the MS-COCO training set, with 60 non-VOC classes as \textit{OOD class}, is used as the unlabeled dataset.
\end{itemize}

\noindent \textbf{Evaluation Metrics.} We use common metrics, including mean Average Precision (mAP) and Average Precision of OOD class (AP$_{u}$), to evaluate performance. Specifically, mAP assesses the performance of ID classes, whereas AP$_{u}$ evaluates the performance of OOD class.

\noindent \textbf{Implementation Details.} Our implementation follows the default settings of Unbiased Teacher \cite{liu2021unbiased}, employing a two-stage training pipeline. The AdamSGD optimizer is utilized with an initial learning rate of 0.01 and a momentum rate of 0.9. Due to GPU limitations, our model is trained on 4 GPUs with a batch size of 16, comprising 8 labeled images and 8 unlabeled images. The memory pool stores embeddings for k+1 classes (k ID classes and 1 OOD class), with a storage capacity $q$ of 256 and a feature dimension $d$ of 128 per class. The IoU threshold $S_{IoU}$ is set to 0.7, and the cosine similarity threshold $S_{cos}$ to 0.5. For $\mathcal{L}_{fc}$, the initial value of $\alpha_{t}$ is 0.1, decaying over iterations. The loss weight $\beta$ for $\mathcal{L}_{uc}$ is set to 1.0.

\subsection{Comparisons with state-of-the-art Methods}
We compare our approach with state-of-the-art OSOD and SSOD methods \cite{sohn2020simple,liu2021unbiased,joseph2021towards,han2022expanding} on COCO-Open-CLS, COCO-Open-SUP, and VOC-COCO benchmarks.

\noindent \textbf{Comparisons on COCO-Open-CLS.} We perform experiments by varying the number of ID classes, as shown in Tab. \ref{tab:coco-open-1}. With fewer ID classes, SSOD models \cite{sohn2020simple,liu2021unbiased} struggle to distinguish OOD class, negatively impacting ID class performance. For instance, with 20 ID classes, UT achieves only 1.50 mAP improvement over the baseline, while our method yields a significant improvement of 5.46 mAP. Moreover, our method achieves 5.56 AP$_{u}$, demonstrating its ability to identify OOD, a capability lacking in both the baseline and UT. 

\noindent \textbf{Comparisons on COCO-Open-SUP.} We conduct experiments by varying the amount of labeled data, as shown in Tab. \ref{tab:coco-open-2}, OSOD models \cite{joseph2021towards,han2022expanding} perform poorly with limited labeled data, while our method remains effective even with small amounts of labeled data, achieving improvements over both the baseline and UT.

\noindent \textbf{Comparisons on VOC-COCO.} Following the satisfactory results achieved on the MS-COCO dataset, we test our method on different datasets. As shown in Tab. \ref{tab:voc-coco}, our method achieves the best results on the combined dataset of PASCAL-VOC and MS-COCO. We improve the performance of UT by 2.67 mAP, with an AP$_{u}$ of 7.86. Compared to the excellent work DCO OSSOD \cite{wang2023online}, our method also achieves an improvement of 1.20 mAP, demonstrating the efficacy of our approach across datasets of varying sizes.

\subsection{Ablation Studies} 
\begin{figure*}
\begin{center}
   \includegraphics[width=0.95\linewidth]{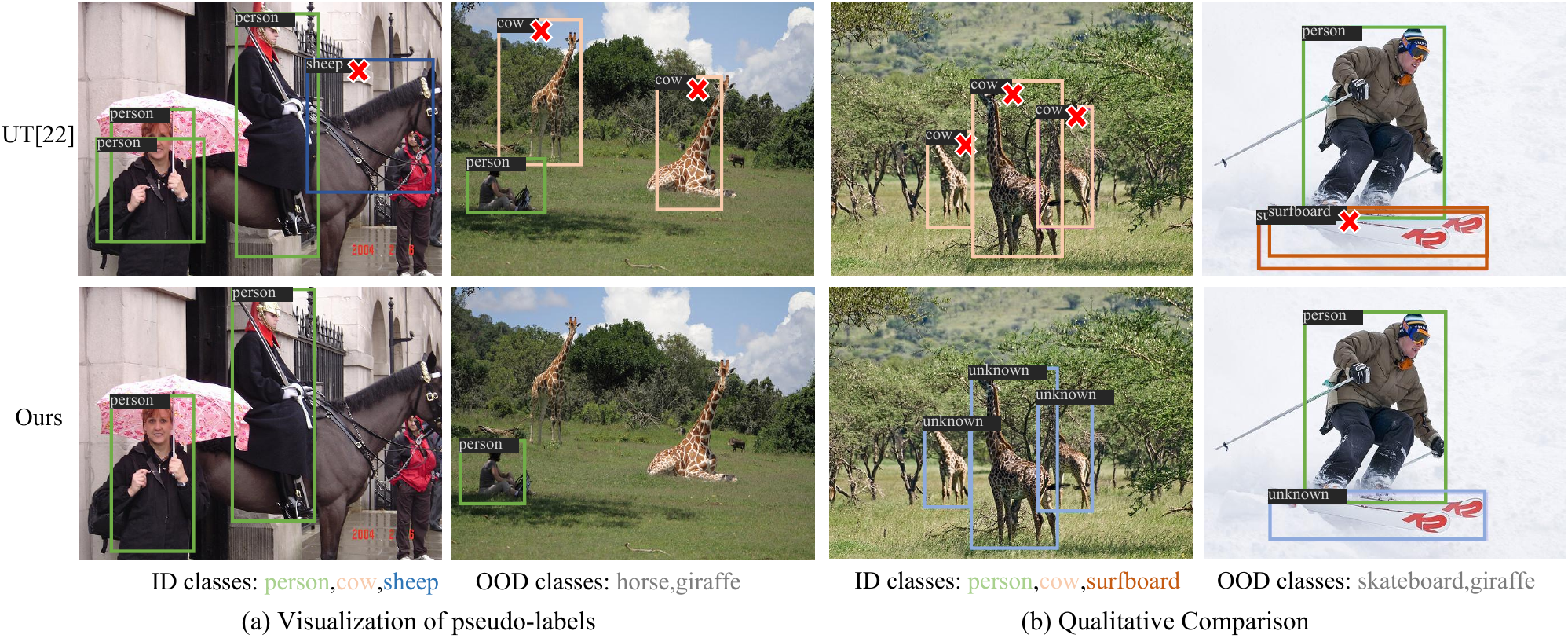}
\end{center}
   \vspace{-1mm}
   \caption{
    (a) Visualization of pseudo-labels from UT (Top) and Ours (Bottom). Our method reduces OOD interference and enhances bounding box quality. (b) Qualitative Comparisons between UT (Top) and Ours (Bottom). UT misclassifies OOD class as ID with high confidence, whereas our method accurately labels them as "unknown".
   }
   \Description{This is an image that shows a visualization of this paper.}
\label{fig:vis}
\vspace{-1mm}
\end{figure*}
\begin{table}[t]
\caption{Effectiveness of $\mathcal{L}_{fc}$ and $\mathcal{L}_{uc}$.}
\vspace{-2mm}
\label{tab:1-abla}
\centering
    \subfloat[Component-wise ablations. 20 ID classes and 4000 labeled images are used for training.
    \label{tab:1-abla_1}]
    { 
    \renewcommand\tabcolsep{4.5pt}
    \renewcommand\arraystretch{1.1} 
    \small
    \begin{tabular}{ll|ll}
    \Xhline{0.7px}
    \centering
     $\mathcal{L}_{fc}$ & $\mathcal{L}_{uc}$ & mAP$_{k}$ & AP$_{u}$  \\
    \hline
    ~\xmark & ~\xmark & 10.75 & 0 \\
    ~\cmark & ~\xmark & 14.59 & 0 \\
    ~\xmark & ~\cmark & 14.21 & 5.14 \\
    \rowcolor{gray!20}
    ~\cmark  & ~\cmark  & \textbf{14.71} & \textbf{5.56} \\
    \Xhline{0.7px}
    \end{tabular}
    }
\centering
    \hspace{2mm}
    \subfloat[Transfer ablations. $\mathcal{L}_{fc}$ and $\mathcal{L}_{uc}$ are applied to STAC. 40 ID classes and 4000 labeled images are used.
    \label{tab:1-abla_2}]
    { 
    \renewcommand\tabcolsep{6.5pt}
    \renewcommand\arraystretch{1.38} 
    \small
    \begin{tabular}{l|ll}
    \Xhline{0.7px}
    \centering
      Methods & mAP$_{k}$ & AP$_{u}$ \\
      \hline
      Label & 15.36 & 0 \\
      STAC & 17.47 & 0 \\
      \rowcolor{gray!20}
      \textbf{STAC+ours} & \textbf{18.38} & \textbf{6.71} \\
      \Xhline{0.7pt}
      \end{tabular}
    }
\vspace{-1mm}
\end{table}
We conduct ablation studies to evaluate the contribution of various designs in our CFL-Detector, as illustrated in Tab. \ref{tab:1-abla}, Tab. \ref{tab:abla-lcon}, and Tab. \ref{tab:abla-luc}. The experiments are conducted on 20 ID classes using 4000 labeled images for training.

\noindent \textbf{Ablations of $\mathcal{L}_{fc}$ and $\mathcal{L}_{uc}$.} 
To assess the impact of $\mathcal{L}_{fc}$ and $\mathcal{L}_{uc}$, we first train the model without these components and progressively integrate each design. As shown in Tab. \ref{tab:1-abla_1}, using only $\mathcal{L}_{fc}$ (2nd result) results in the model's inability to identify the OOD class. However, with the introduction of $\mathcal{L}_{uc}$ (3rd result), AP$_u$ increases from 0 to 5.14, demonstrating that $\mathcal{L}_{uc}$ enables the model to learn uncertainty probabilities. The combined use of both $\mathcal{L}_{fc}$ and $\mathcal{L}_{uc}$ yields the best performance (4th result). 

Furthermore, $\mathcal{L}_{fc}$ and $\mathcal{L}_{uc}$ can be seamlessly integrated into other SSOD approaches. As shown in Tab. \ref{tab:1-abla_2}, applying them to the STAC \cite{sohn2020simple} increases mAP from 17.47 to 18.38 and enables OOD class identification, achieving 6.71 AP$_{u}$. This demonstrates the effectiveness of our approach across different SSOD architectures.

\begin{table}
\caption{Ablations on different settings of $\mathcal{L}_{fc}$.}
\vspace{-2mm}
\label{tab:abla-lcon}
    \subfloat[Influences on the storage of OOD class.
    \label{tab:con-abla_1}]
    { 
    \centering
    \renewcommand\tabcolsep{8pt}
    \renewcommand\arraystretch{1.45} 
    \small
    \begin{tabular}
    {l|ll}
    \Xhline{0.7px}
    \centering
      & ~\xmark & \cellcolor{gray!20} ~\cmark  \\
    \hline
    mAP$_{k}$ & 14.3 & \cellcolor{gray!20} \textbf{14.7}  \\
    AP$_{u}$ & 4.9 & \cellcolor{gray!20} \textbf{5.6}  \\ 
    \Xhline{0.7px}
    \end{tabular}
    }
    \hspace{2mm}
    \subfloat[Different choices of storage size ($q$) and feature dimensions ($dim$).
    \label{tab:con-abla_2}]
    { 
    \centering%
    \small 
    \renewcommand\tabcolsep{4pt}
    \renewcommand\arraystretch{1.1} 
        \begin{tabular}{l|lllll}
        \Xhline{0.7px}
        \centering
        $q$ & 128 & \cellcolor{gray!20} 256 & 512 & 256  \\
	$dim$ & 128 & \cellcolor{gray!20} 128 & 128 & 256  \\ \hline
	mAP$_k$ & 14.5 & \cellcolor{gray!20} 14.7 & 14.5 & \textbf{14.9}    \\
	AP$_{u}$ & 5.6 & \cellcolor{gray!20} 5.6 & \textbf{5.7} & 5.2     \\
        \Xhline{0.7px}
        \end{tabular}
    }
\vspace{2mm}

    \subfloat[Experimental results on different loss weight settings.
    \label{tab:con-abla_3}]
    { 
    \centering
    \renewcommand\tabcolsep{4.5pt}
    \renewcommand\arraystretch{1.45} 
    \small
    \begin{tabular}
    {l|lll}
    \Xhline{0.7px}
    \centering
     $\alpha_{t}$ & 0.01 & \cellcolor{gray!20} 0.1 & 0.5  \\
    \hline 
    mAP$_{k}$ & \textbf{14.8} & \cellcolor{gray!20} 14.7 & 14.2 \\
    AP$_{u}$ & 5.2 & \cellcolor{gray!20} 5.6 & \textbf{6.0} \\
    \Xhline{0.7px}
    \end{tabular}
    }
    \hspace{2mm}
    \subfloat[Different choices of $S_{IoU}$ and $S_{cos}$ threshold.
    \label{tab:con-abla_4}]
    { 
    \centering%
    \small
    \renewcommand\tabcolsep{4pt}
    \renewcommand\arraystretch{1.1} 

        \begin{tabular}{l|llll}
        \Xhline{0.7px}
        \centering
        $S_{IoU}$ & 0.5 & 0.5 & \cellcolor{gray!20} 0.7 & 0.7  \\
	$S_{cos}$ & 0.5 & 0.7 & \cellcolor{gray!20} 0.5 & 0.7  \\ \hline
	mAP$_k$ & \textbf{14.8} & 14.6 & \cellcolor{gray!20} 14.7 & 14.6  \\
	AP$_{u}$ & 5.4 & 5.6 & \cellcolor{gray!20} \textbf{5.6} & 5.5   \\
        \Xhline{0.7px}
        \end{tabular}
    }

\end{table}
\noindent \textbf{Ablations of various choices in $\mathcal{L}_{fc}$.} The effectiveness of $\mathcal{L}_{fc}$ depends on the design of the memory pool. As illustrated in Tab. \ref{tab:con-abla_1}, storing additional OOD class features enriches the semantic information within the memory pool, enhancing the model's ability to distinguish between different classes. Secondly, as shown in Tab. \ref{tab:con-abla_2}, a feature storage size of 256 (with a dimension of 128) achieves optimal results, indicating the importance of balancing storage capacity ($q$) and feature dimension ($dim$) to preserve diversity. Furthermore, Tab. \ref{tab:con-abla_4} indicates that a higher IoU threshold ($S_{IoU}$) ensures that only high-quality features with better spatial alignment with ground truth are retained, while cosine similarity ($S_{cos}$) guarantees robust semantic information, collectively improving the memory pool's accuracy. Lastly, Tab. \ref{tab:con-abla_3} highlights the importance of appropriately adjusting the weight ($\alpha_t$), as overly high or low values degrade performance, with 0.1 being the best choice. 

\begin{table}
\caption{Ablations on different settings of $\mathcal{L}_{uc}$.}
\label{tab:abla-luc}
\vspace{-2mm}
\centering
    \hspace{-2mm} 
    \subfloat[Effectiveness of $\mathcal{L}_{uc}$, composed of $w_k\mathcal{L}_{ce}$ and $w_u\mathcal{L}_{ce}$.
    \label{tab:uc-abla_1}]
    { 
    \renewcommand\tabcolsep{2.5pt}
    \renewcommand\arraystretch{1.1} 
    \small
    \begin{tabular}{cc|cc}
    \Xhline{0.7px}
    \centering
    $w_k\mathcal{L}_{ce}$ & $w_u\mathcal{L}_{ce}$ & mAP$_{k}$ & AP$_{u}$  \\
    \hline
    ~\xmark & ~\xmark & 10.75 & 0 \\
    ~\cmark & ~\cmark & 14.17 & 2.99 \\ 
    \rowcolor{gray!20}
    ~\xmark & ~\cmark & \textbf{14.71} & \textbf{5.56} \\
    \Xhline{0.7px}
    \end{tabular}
    }
    \hspace{1mm} 
    \subfloat[Experimental results on different $\mathcal{L}_{uc}$ weight.
    \label{tab:uc-abla_2}]
    { 
    \renewcommand\tabcolsep{10.5pt}
    \renewcommand\arraystretch{1.1} 
    \small
    \begin{tabular}{l|ll}
    \Xhline{0.7px}
    \centering
     $\beta$ & mAP$_{k}$ & AP$_{u}$  \\
    \hline 
    0.5 & 15.01 & 4.55 \\
    0.8 & 14.88 & 5.11 \\
    \rowcolor{gray!20}
    \textbf{1.0} & \textbf{14.71} & \textbf{5.56} \\
    \Xhline{0.7px}
    \end{tabular}
    }
\end{table}
\noindent \textbf{Ablations of various choices in $\mathcal{L}_{uc}$.} 
\label{sec:Effect of Uncertainty classification loss}
Tab. \ref{tab:uc-abla_1} highlights the improvement in OOD class identification achieved by integrating $\mathcal{L}_{uc}$. However, joint training with $w_k\mathcal{L}_{ce}$ during the semi-supervised stage introduces pseudo-label instability, reducing the accuracy for both ID and OOD classes. To address this, we optimize only $w_u\mathcal{L}_{ce}$ during the semi-supervised stage. We also conduct an ablation study on the loss weight ($\beta$). As shown in Tab. \ref{tab:uc-abla_2}, lower weight improves ID performance but significantly degrade OOD performance. The best results are obtained with a weight of 1.0.

\subsection{Qualitative Results} 
\label{sec:Qualitative Results}
The quality of pseudo-labels is crucial to model performance. Fig. \ref{fig:vis} (a) presents visualizations of pseudo-labels for UT and our method. UT misidentifies OOD class as ID classes and produces low-quality bounding boxes with overlapping regions. In contrast, our model resolves these issues and generates more accurate pseudo-labels.

In Fig. \ref{fig:vis} (b), we compare the qualitative results of UT with our method. UT often misclassifies OOD objects as ID classes with high confidence, while our method correctly labels such objects as "unknown". Additionally, our method provides more accurate bounding box coordinates and eliminates the problem of predicting multiple boxes for the same object, which is common in UT.

\section{Conclusion}
This paper presents a novel OSSOD framework that addresses challenging tasks by optimizing two loss functions: $\mathcal{L}_{fc}$ and $\mathcal{L}_{uc}$. $\mathcal{L}_{fc}$ enhances class separation in the feature space, while $\mathcal{L}_{uc}$ enables effective identification of the OOD class. Extensive experiments show that our method achieves state-of-the-art performance across various datasets and can be seamlessly integrated into other SSOD methods. Additionally, ablation studies validate the effectiveness of each component and the selection of optimal hyper-parameters.

\noindent \textbf{Limitation.} 
While our approach improves OOD class accuracy, it may cause a margin decline in ID class performance, suggesting a need for further research to balance this trade-off.

\noindent \textbf{Acknowledgment.}
This research was funded by the Fundamental Research Funds for the Central Universities (2024XKRC082) and the National NSF of China (No.U23A20314).

\bibliographystyle{ACM-Reference-Format}
\bibliography{sample-base.bib}

\end{document}